\documentclass[journal]{IEEEtran}

\ifCLASSINFOpdf
\else
   \usepackage[dvips]{graphicx}
\fi
\usepackage{url}
\usepackage{amssymb}
\usepackage{booktabs}
\usepackage{graphicx}
\usepackage{amsmath}
\usepackage{booktabs}
\usepackage{multirow}
\usepackage[table]{xcolor}
\usepackage{xcolor}
\usepackage[maxnames=99]{biblatex}
\usepackage[breaklinks,colorlinks]{hyperref}

\begin{document}

\title{Generalized Portrait Quality Assessment}


\author{Nicolas Chahine, Sira Ferradans, Javier Vazquez-Corral, Jean Ponce
\thanks{
\scriptsize{This work was funded in part by the French government under the management of Agence Nationale de la Recherche as part of the “Investissements d’avenir” program, reference ANR-19-P3IA-0001 (PRAIRIE 3IA Institute), the Louis Vuitton/ENS chair in artificial intelligence and the Inria/NYU collaboration.  This work was performed using HPC resources from GENCI-IDRIS (Grant 2023-AD011013850). NC was supported in part by a DXOMARK/PRAIRIE CIFRE Fellowship. JVC was supported  by Grant PID2021-128178OB-I00 funded by MCIN/AEI/10.13039/501100011033, ERDF "A way of making Europe", the Departament de Recerca i Universitats from Generalitat de Catalunya with reference 2021SGR01499.}}
\thanks{\scriptsize{N. Chahine is with DXOMARK, Paris, France \& the Departement d’informatique de l’Ecole normale suprerieure (ENS-PSL, CNRS, Inria) (e-mail: nchahine@dxomark.com).}}
\thanks{\scriptsize{ Ferradans is with DXOMARK, Paris, France (e-mail: sferradans@dxomark.com).}}
\thanks{\scriptsize{J. Vazquez Corral, is with Computer Vision Center, Barcelona, Spain \& Computer Sciences Departament Universitat Autònoma de Barcelona, Barcelona, Spain (e-mail: javier.vazquez@cvc.uab.cat).}}
\thanks{\scriptsize{J. Ponce is with the Departement d’informatique de l’Ecole normale suprerieure (ENS-PSL, CNRS, Inria) \& the Institute of Mathematical Sciences and Center for Data Science, New York University (e-mail: jean.ponce@ens.fr).}}}
\maketitle

\begin{abstract}
Automated and robust portrait quality assessment (PQA) is of paramount importance in high-impact applications such as smartphone photography. This paper presents FHIQA, a learning-based approach to PQA that introduces a simple but effective quality score rescaling method based on image semantics, to enhance the precision of fine-grained image quality metrics while ensuring robust generalization to various scene settings beyond the training dataset. The proposed approach is validated by extensive experiments on the PIQ23 benchmark and comparisons with the current state of the art. The source code of FHIQA will be made publicly available on the PIQ23 GitHub repository at \url{https://github.com/DXOMARK-Research/PIQ2023}.


\end{abstract}


\IEEEpeerreviewmaketitle

\section{Introduction}

\IEEEPARstart{S}{martphones} have significantly altered the landscape of photographic practices, with a notable emphasis on portrait photography. As the consumer base becomes increasingly discerning, there is a clear escalation in expectations regarding the quality of portrait images. This trend pushes a corresponding advancement in camera technology driving manufacturers to focus on improving image quality, which usually involves developing costly image quality assessment (IQA) protocols for optimizing camera performance. Consequently, automated IQA methods are employed to cut the cost of smartphone camera tuning. 

Traditional IQA \cite{larson2010most, ponomarenko2015image, ponomarenko2009tid2008, sheikh2006statistical} offers relevant insights about image quality but often doesn't capture the complexities of modern camera systems and weakly correlates with human perception \cite{fang2020perceptual}. Deep learning-based IQA has emerged as a promising alternative leveraging the widespread availability of modern smartphone images and image quality datasets. Popular datasets like LIVE \cite{sheikh2006statistical}, CSIQ \cite{larson2010most}, and TID \cite{ponomarenko2015image, ponomarenko2009tid2008} present a large amount of synthetically distorted images, but they don't encapsulate the multifaceted nature of modern smartphone camera systems. ``In-the-wild" collections, including LIVE Challenge \cite{ghadiyaram2015massive}, KonIQ-10k \cite{hosu2020koniq}, and PaQ-2-PiQ \cite{ying2020patches}, offer a better representation of real-world conditions, but their uncontrolled data collection and subjective labeling make them less suited for rigorous digital camera evaluations. Recently, datasets such as PIQ23 \cite{chahine2023image} have emerged, offering a setting tailored to portrait photography including diverse scenes, each independently annotated by image quality experts.
The PIQ23 dataset is crafted to achieve high precision in annotations by employing two key strategies: a) utilizing pairwise comparisons for image annotation, a method known for enhancing consistency in IQA experiments \cite{mantiuk2012comparison, perez2019pairwise}; and b) categorizing images by content into distinct scenes to harness the human visual system's precision when evaluating images with shared content. Notably, the PIQ23 dataset distinguishes itself with a comprehensive array of portrait images sourced from over 100 smartphone models, encompassing 50 varied scenarios that represent a wide range of photographic conditions.

\begin{figure}
\centerline{\includegraphics[width=0.93\columnwidth]{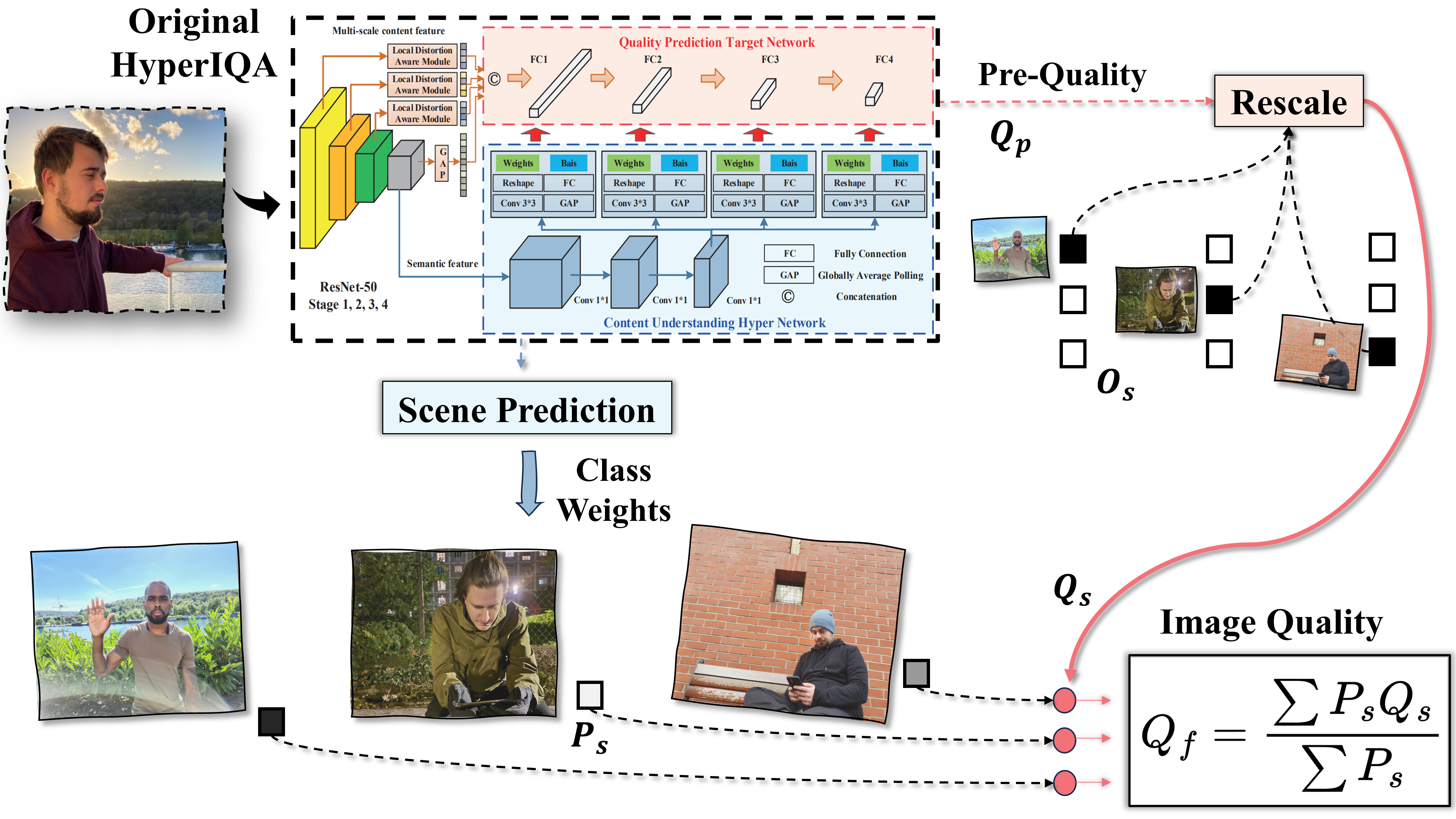}}
\caption{Diagram of FULL-HyperIQA (FHIQA). The figure illustrates how FHIQA processes input images, extracts semantic information, and adapts the quality prediction based on scene-specific evaluations.}
\label{fig:FULL-HyperIQA}
\end{figure}

Blind IQA (BIQA) has gained significant interest in recent years, as it offers universal metrics for quality assessment without the need for pristine reference images, often scarce in practical photography scenarios. BIQA methods, based on convolutional neural networks (CNN) \cite{kang2014convolutional, kim2016fully, zhang2018blind}, have shown marked improvements over their classical counterparts, due to their ability to extract perceptual quality information \cite{ghadiyaram2017perceptual, mittal2012no, moorthy2011blind, saad2012blind,  ye2012unsupervised}.  
However, many existing BIQA methods exhibit limitations in capturing scene-specific semantics, often treating diverse scenes with a one-size-fits-all approach. 
Given that image quality is inherently subjective and varies under different conditions, numerous studies underscore the significance of integrating semantic information into the assessment of image quality, tackling a challenge known as domain shift \cite{sun2021blind, zerman2017extensive, zhang2020learning, zhang2021uncertainty}. This is often achieved through a semantics-aware multitasking framework \cite{chahine2023image, fang2020perceptual, feng2023learning, huang2020multi, su2020blindly}.
Moreover, the variability of quality scales across IQA datasets introduces ambiguities, complicating the aggregation of IQA insights. As a result, most BIQA methods struggle to generalize to new conditions, underscoring a critical challenge in cross-domain image quality evaluations.


In tackling the limitations of current BIQA methodologies, such as combining IQA knowledge and limitations on generalization, this work presents FULL-HyperIQA (FHIQA), an advancement over prior HyperNetwork models (\cite{su2020blindly, chahine2023image}). FHIQA not only considers scene-specific semantics in predicting image quality but also merges knowledge across various scenes for improved generalization. This approach offers a context-aware IQA method, better suited to adapting to new, unfamiliar conditions.

Our contribution is a new BIQA model, FHIQA, focusing on, a) precision in quality metrics, meaning, the capacity to generate highly granular results thus enabling the differentiation of closely matched cameras; and b) generality by extending the scope of previous work to tackle image content (scenes) not encountered during training. Our comprehensive tests on a newly introduced PIQ23 test split demonstrate FHIQA's enhanced generalization capabilities over existing BIQA benchmarks.

\section{Method}

We present FULL-HyperIQA (Fig. \ref{fig:FULL-HyperIQA}), an enhanced version of SEM-HyperIQA \cite{chahine2023image}, with a specific focus on adapting to scenes not encountered during training. 
\paragraph{Scene-specific rescaling} SEM-HyperIQA combines the HyperIQA architecture \cite{su2020blindly}, which integrates semantic information, with multitasking, which allows scene-specific quality score rescaling. Semantic features from multiple random crops are concatenated and fed to a multi-layer perceptron (MLP) that predicts the scene category for the image ($s$). Then, the predicted category is passed to a smaller MLP that predicts a multiplier $a^s_i$ and offset $b^s_i$ to adapt the predicted quality score of each patch, $q_i$, to its respective scene quality scale, such as $\hat{q}^s_i = a^s_i q_i + b^s_i$, where $\hat{q}^s_i$ is the rescaled quality score of patch $i$. The final image quality score is computed by averaging the individual patch scores. This approach provides a good basis for solving the problem of domain shift and scene-specific rescaling. However, SEM-HyperIQA does not explicitly allow generalization to new scenes, since it relies on predicting a single category for each image, supposing we only use images from known 
scenes.

\paragraph{Understanding new content} The novelty of FHIQA lies in its approach to quality score rescaling. Instead of adjusting the final score based solely on the scene predicted for the image, FHIQA utilizes the entire scene prediction vector. This vector can be interpreted as the set of coordinates in the vectorial space of scenes defined in the training set. The coordinate (or weight) of each scene indicates how similar the input is to that scene. For the scenes of the training set, the input image is assigned a one-hot vector corresponding to that scene ($O_s$ in Fig. \ref{fig:FULL-HyperIQA}). During training, the model is trained to predict the scene in the image and to align the predicted pre-quality score ($Q_p \in \mathbb{R}$ in Fig. \ref{fig:FULL-HyperIQA}) with the scene scale, by passing the predicted classification vector as input to the rescaling layer. The only difference is that we split the predicted classification vector into separate one-hot encodings and then rescale the pre-quality score on each of the training scenes. The final quality score is obtained by weighting the rescaled scores with the predicted classification vector. Finally, our main concern is to generalize the quality predictions to scene categories that were not included in the training set.
We hypothesize that the information needed for this is naturally encoded in the class prediction weights. In short, if the scene classification vector considers an input image to be closer to one scene category, its quality score should also be closer to that category. Mathematically, we can define our hypothesis as follows:

\begin{equation}
Q_f = \frac{\sum_{i=1}^{k} P_{s_i} (a^s_{i} Q_p + b^s_{i}) }{\sum_{j=1}^{k} P_{s_j}}\text{, where}
\end{equation}

\begin{itemize}
    \item \( Q_f \) is the final quality score for the input image.
    \item \( P_{s_i} \) denotes the weight that the input image belongs to scene \( s_i \).
    \item $a^s_{i}$ and $b^s_{i}$ are the multiplier and offset, respectively, for the scene \( s_i \).
    \item $Q_p$ is the pre-quality score predicted by the fully connected layer.
\end{itemize}
To optimize this approach, instead of considering all scenes, only the top $k$ scenes are considered. The weights are then normalized, and a weighted average is taken based on these top $k$ scenes to produce the final quality score for the input image.

\section{Dataset}

\begin{figure*}[!h]
\centering
\includegraphics[width=0.8\textwidth]{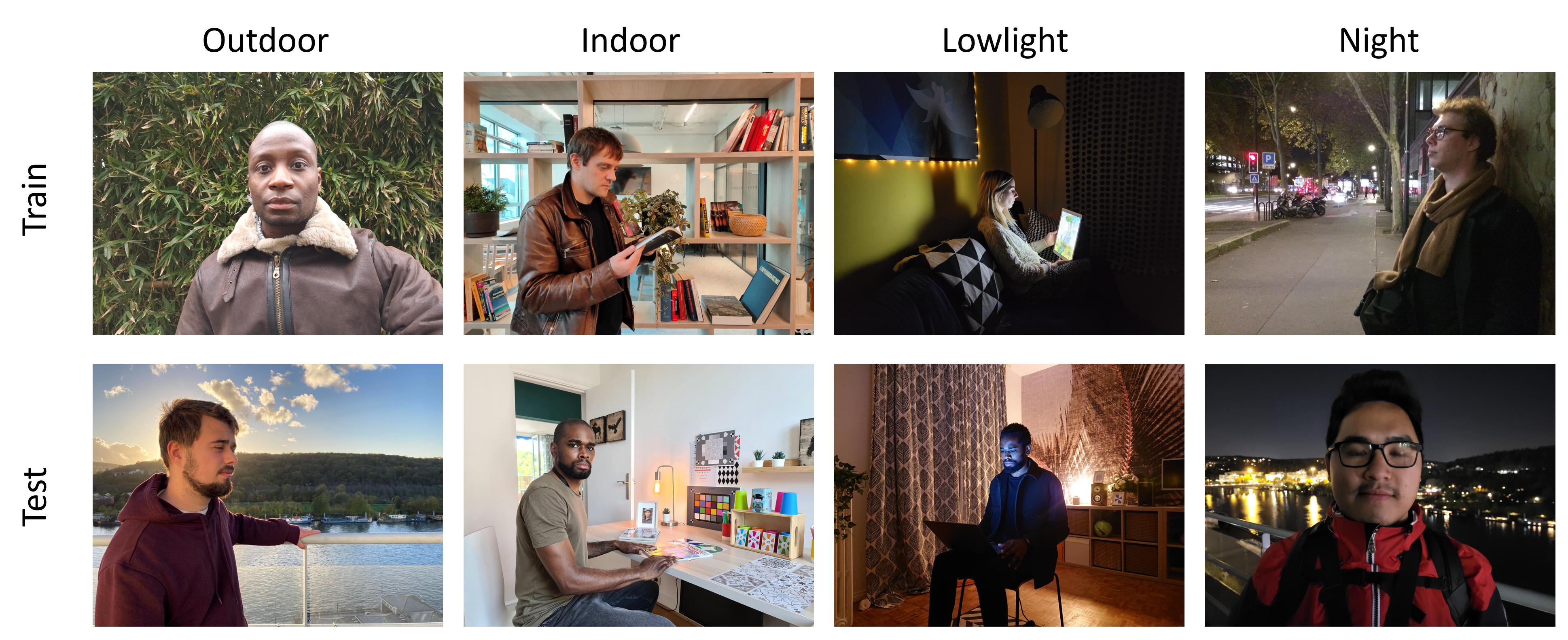}
\setlength{\tabcolsep}{2pt}

\caption{Examples from the new scene split for PIQ23. The test set incorporates various framing settings, backgrounds, subject characteristics, and weather conditions that are significantly distinct from the training set.}
\label{tab:piq23}
\end{figure*}

Since our work focuses on portrait quality assessment, we have chosen to evaluate our model on PIQ23 \cite{chahine2023image}. Portrait photography captures the essence of human emotions, expressions, and individual characteristics, presenting unique challenges in the domain of image quality assessment. Recognizing the need for a specialized benchmarking tool, PIQ23 \cite{chahine2023image} has been introduced as a dataset aimed at a more subtle evaluation of IQA models within the context of portrait photography. Benchmarking against PIQ23 ensures a comprehensive and representative evaluation of real-world scenarios for smartphone portrait photography. A significant challenge in IQA is the model's ability to generalize to multiple scenes, especially those not encountered during training. To rigorously evaluate this aspect, we have carefully chosen 15 out of 50 scenes featured in PIQ23 for testing, and the rest for training. This selection accounts for approximately 30\% of the total images, uniformly distributed across the different lighting conditions, encompassing around 1486 out of the 5116 images of PIQ23. During the scene selection process, we ensured that both sets captured a rich blend of conditions, i.e., framing, lighting, and skin tones (Fig. \ref{tab:piq23}).

\section{Experiments}

\subsection{Baselines methods}
We have compared FHIQA with several well-known BIQA models: DB-CNN \cite{zhang2018blind}, HyperIQA \cite{su2020blindly}, MUSIQ \cite{ke2021musiq}, and SEM-HyperIQA \cite{chahine2023image}. Using their official implementations, we have fine-tuned these models on the PIQ23 dataset. DB-CNN and two MUSIQ models were initially trained on the LIVE Challenge, KonIQ-10k, and PaQ-2-PiQ datasets, respectively. For all HyperIQA variants, only the Resnet50 backbone was pre-trained on ImageNet without any subsequent IQA pre-training. Due to HyperIQA's input size constraint of 224x224, we had to modify its architecture to handle resolutions that are multiples of 224 to be able to train on larger images.

\subsection{Training strategy}
We test different training configurations for all the proposed methods and report the best results. Specifically, we randomly crop the images to square patches of one of the three following sizes: 224 (5 patches per image), 672 (3 patches), and 1344 (1 patch). 
We use Adam stochastic optimization with different learning rates between $10^{-6}$ and $10^{-4}$. For HyperIQA, SEM-HyperIQA, and FHIQA, we adopt different learning rates per module. For instance, we apply a smaller learning rate for the Resnet50 backbone compared to the hypernetwork, the rescaling, and the classification blocks. We fix the training for 300 epochs and adopt a learning rate decay factor of 0.05 for every 10 epochs. For FHIQA, we experimented with different values of $k$ (the number of scenes utilized), including 3, 5, 10, and 25. We use early stopping with a patience of 40 epochs. Finally, we use Huber loss for the quality output and cross-entropy loss for the classification output of multitasking models. For the latter models, we apply a weighted sum of losses with a weight of 0.5 for the classification loss and 1.0 for the quality loss. 

\subsection{Metrics}

To evaluate the performance, we compute Pearson's linear correlation coefficient (PLCC), Spearman’s rank correlation coefficient (SRCC), Kendal's rank correlation (KRCC), the averaged correlations, and the mean absolute error (MAE) between the model outputs and the ground-truth scores. 
In the PIQ23 dataset, each scene is annotated individually, thus quality scores cannot be merged. Therefore, we calculate metrics for each scene separately. The aggregate performance across all scenes for a given metric is determined by the median, $M_{\text{Med}} = M_{(\tfrac{s}{2})}$
where \( s \) denotes the total number of scenes, and \( M_{(i)} \) represents the \( i \)-th smallest scene metric value among the sorted scenes. For the early stopping, we evaluate our models on their SRCC performance. It is noteworthy that achieving a high SRCC doesn't necessarily correlate with other metrics. For future works, we might indeed benefit from considering a mix of metrics for early stopping, as small variations in correlation might arise due to a variety of factors, such as the number of images for each scene, score distribution, etc.


\subsection{Results}
\begin{table*}[t]
\centering
\caption{
Performance metrics of various IQA models on PIQ23. The results are presented as the MEDIAN across scenes for each metric. 
}
\setlength{\tabcolsep}{3pt}
\resizebox{0.77\textwidth}{!}{
\begin{tabular}{l|cccc|cccc|cccc}
\toprule
\multirow{3}{*}{Model\textbackslash Attribute} & \multicolumn{4}{c|}{Overall} & \multicolumn{4}{c|}{Exposure} & \multicolumn{4}{c}{Details} \\
\cmidrule{2-13}
 & SRCC & PLCC & KRCC & MAE & SRCC & PLCC & KRCC & MAE & SRCC & PLCC & KRCC & MAE \\
\midrule
DB-CNN (LIVE C) & 0.59 & 0.64 & 0.43 & {1.04} & 0.69 & 0.69 & 0.51 & {0.91} & 0.59 & 0.51 & 0.45 & {0.99} \\
MUSIQ (KonIQ-10k) & \underline{0.76} & \underline{0.75} &\underline{0.57} & \textbf{0.95} & 0.74 & 0.70 & 0.55 & 0.93 & 0.71 & 0.67 & 0.52 & 0.88 \\
MUSIQ (PaQ-2-PiQ) & 0.74 & 0.74 & 0.54 & 1.09 & \textbf{0.79} & \textbf{0.78} & \textbf{0.59} & 0.87 & 0.72 & \textbf{0.77} & \underline{0.53} & 0.90 \\
HyperIQA* & 0.74 & 0.74 & 0.55 & \underline{0.99} & 0.69 & 0.68 & 0.50 & \underline{0.86} & 0.70 & 0.67 & 0.50 & 0.94 \\
SEM-HyperIQA* & {0.75} & \underline{0.75} & {0.56} & {1.03} & {0.72} & 0.70 & {0.53} & 0.97 & 0.73 & 0.65 & \textbf{0.55} & 0.88 \\
SEM-HyperIQA-CO* & 0.74 & 0.74 & 0.55 & 1.04 & 0.70 & 0.70 & 0.52 & 0.94 & \textbf{0.75} & {0.71} & \textbf{0.55} & \underline{0.85} \\
\rowcolor{gray!25}
\textbf{FULL-HyperIQA}* & \textbf{0.78} & \textbf{0.78} & \textbf{0.59} & 1.12 & \underline{0.76} & \underline{0.71} & \underline{0.57} & \textbf{0.85} & \underline{0.74} & \underline{0.72} & \textbf{0.55} & \textbf{0.80} \\
\bottomrule
\multicolumn{13}{p{270pt}}{*ImageNet, backbone only; \textbf{best}; \underline{second best}.}\\
\end{tabular}
}

\label{tab:benchmarks}
\end{table*}

\begin{figure*} 
    \centering
    \begin{tabular}{ccc}
        {\includegraphics[width=0.66\columnwidth]{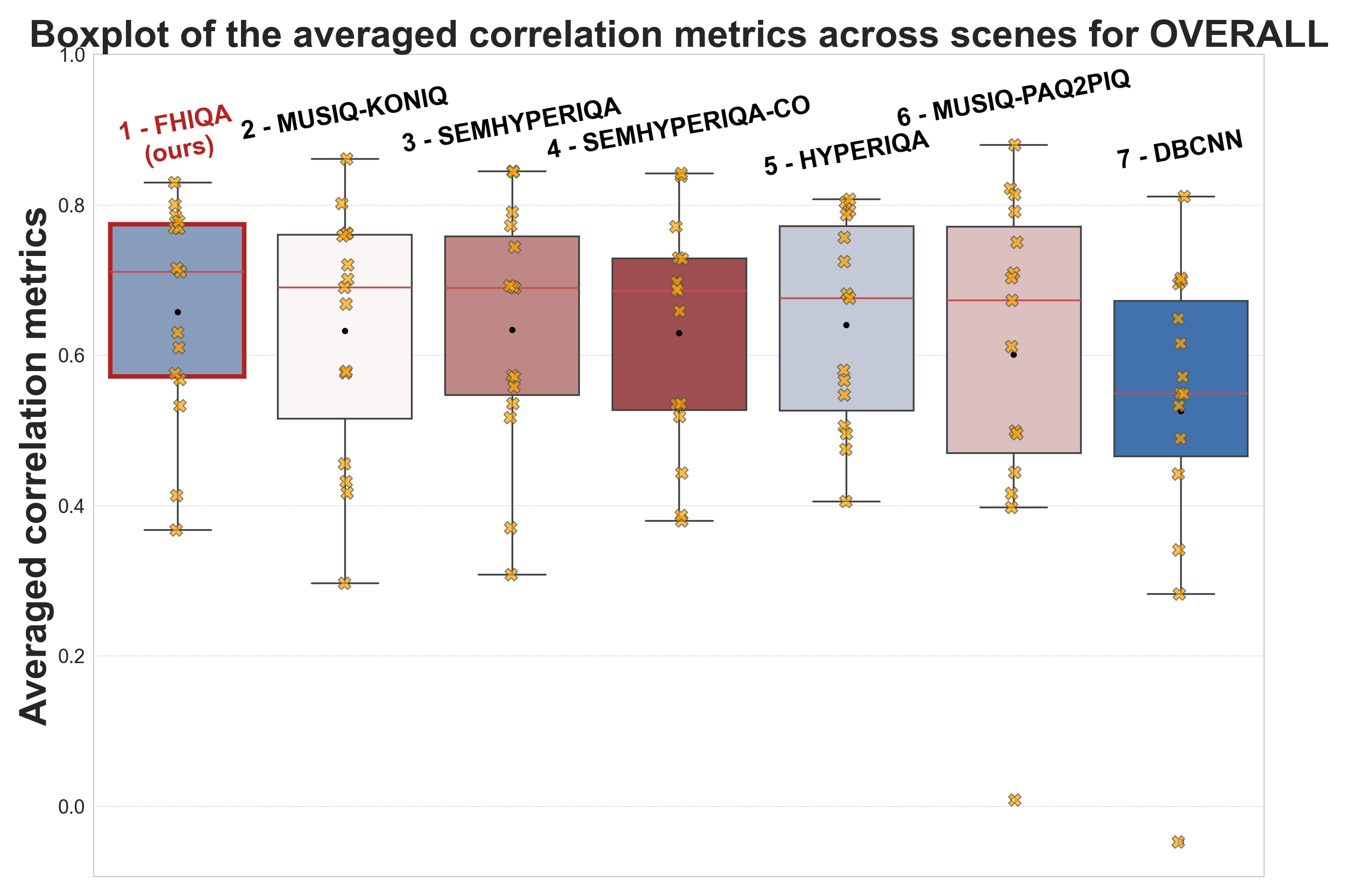}}&
        \includegraphics[width=0.66\columnwidth]{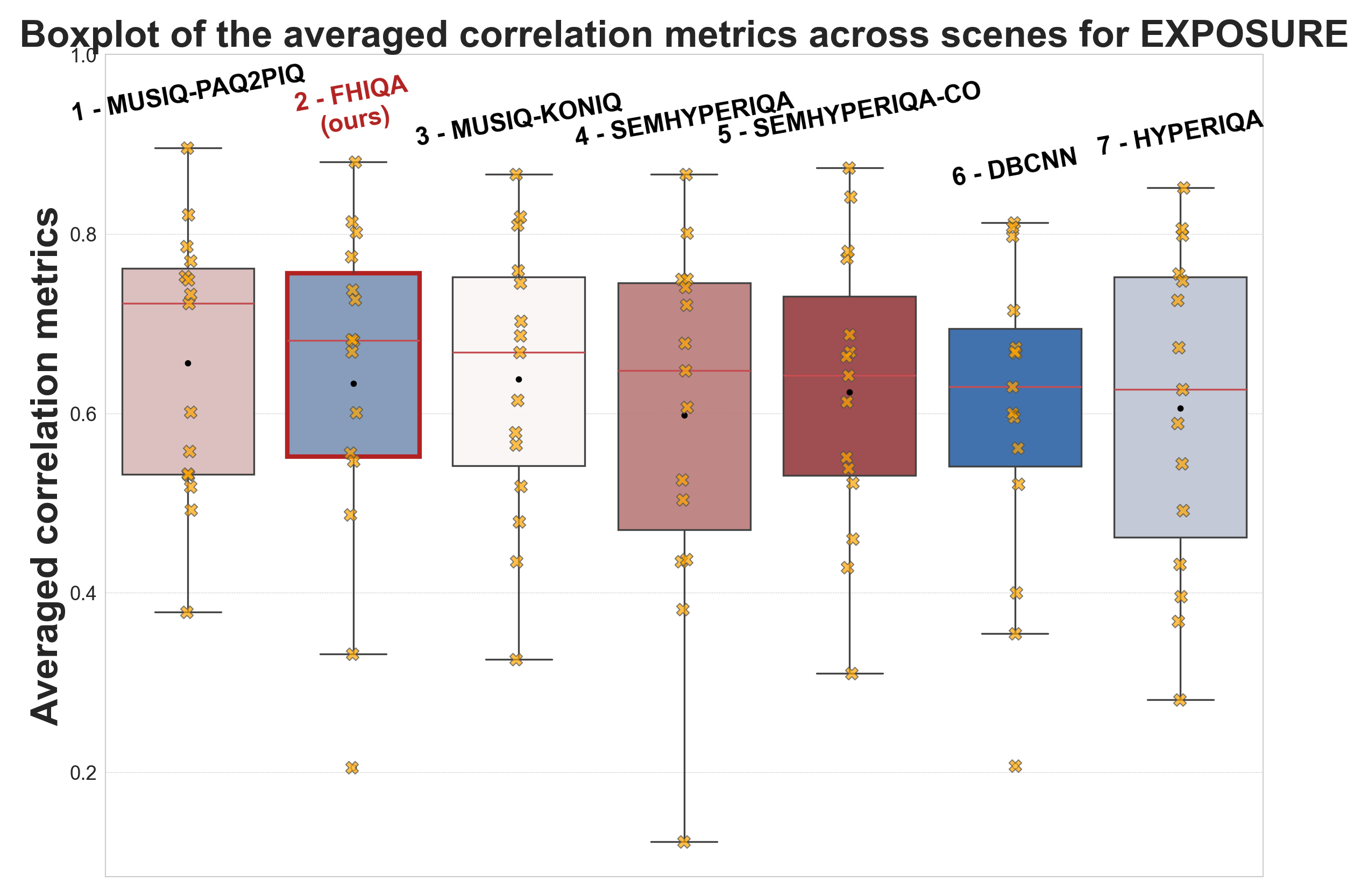} &
        \includegraphics[width=0.66\columnwidth]{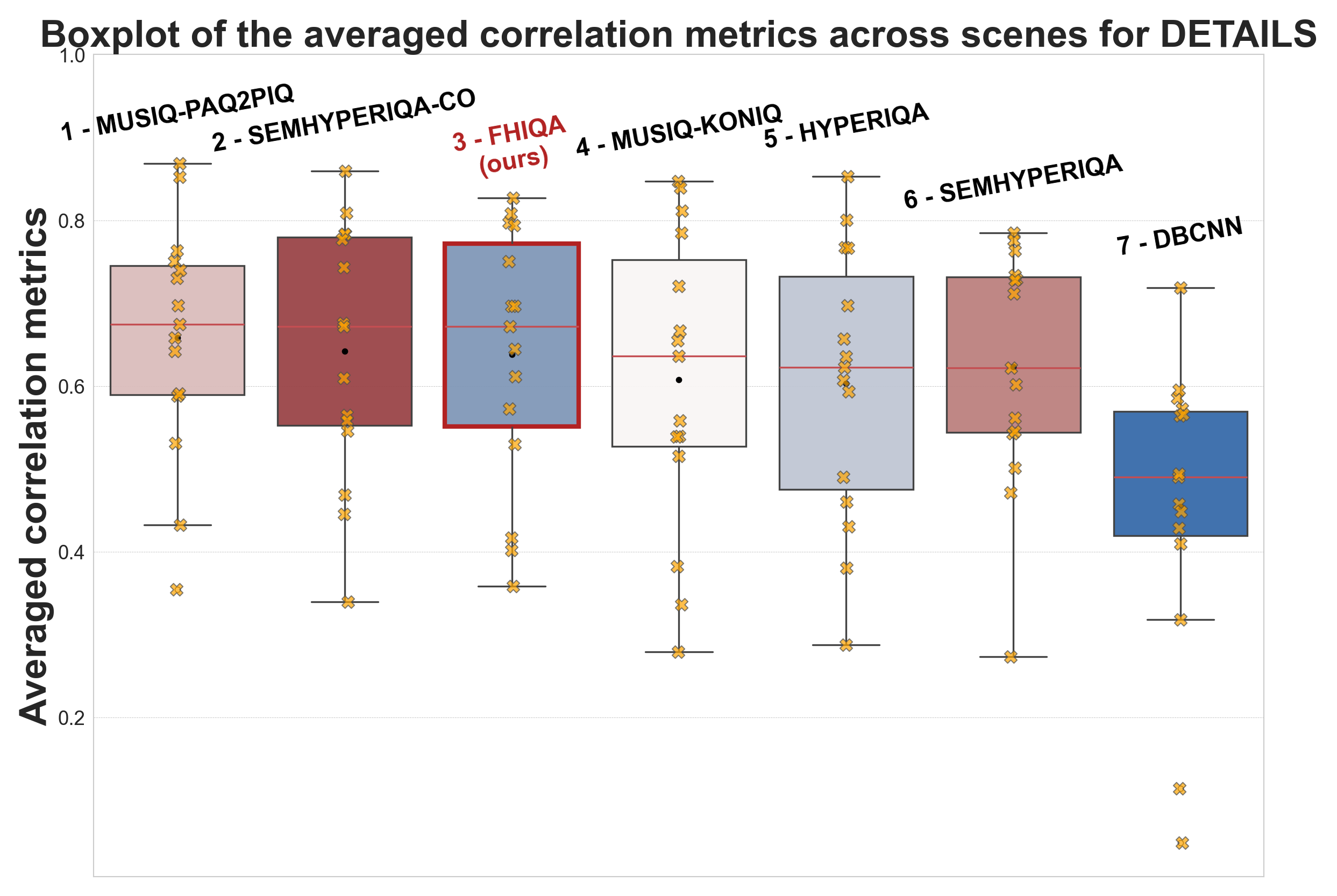} \\
    \end{tabular}
    \caption{Comparative analysis of IQA models based on the averaged correlation metrics distribution across all scenes and for the three attributes of PIQ23.}
    \label{fig:boxplots}
\end{figure*}

\begin{figure*} 
    \centering
    \begin{tabular}{cc}
        \includegraphics[width=\columnwidth]{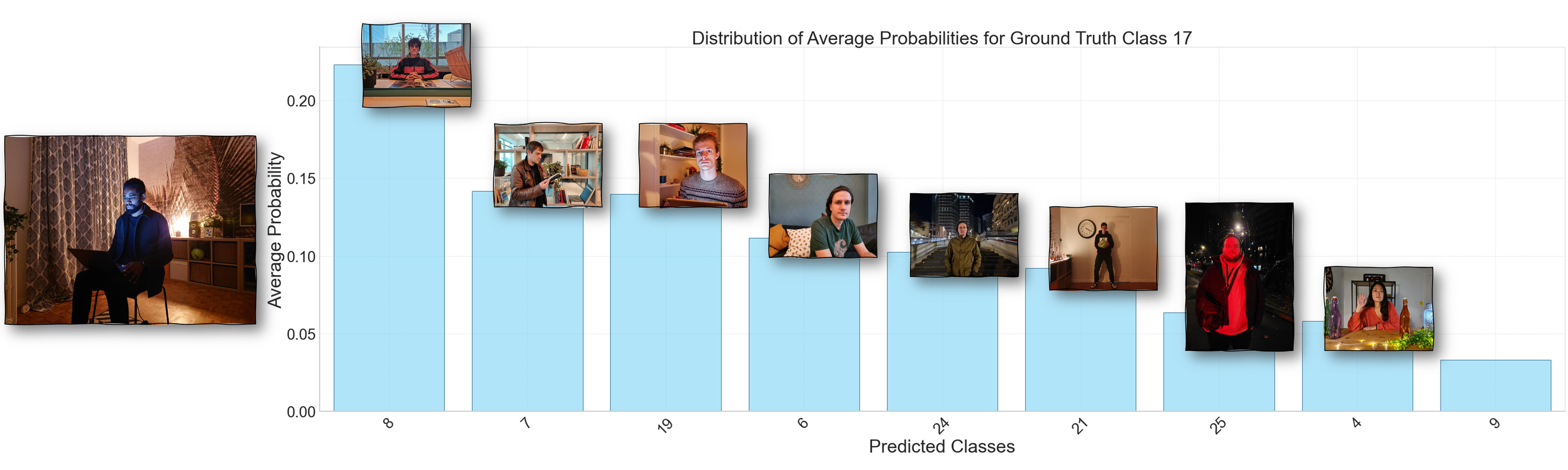} &
        \includegraphics[width=\columnwidth]{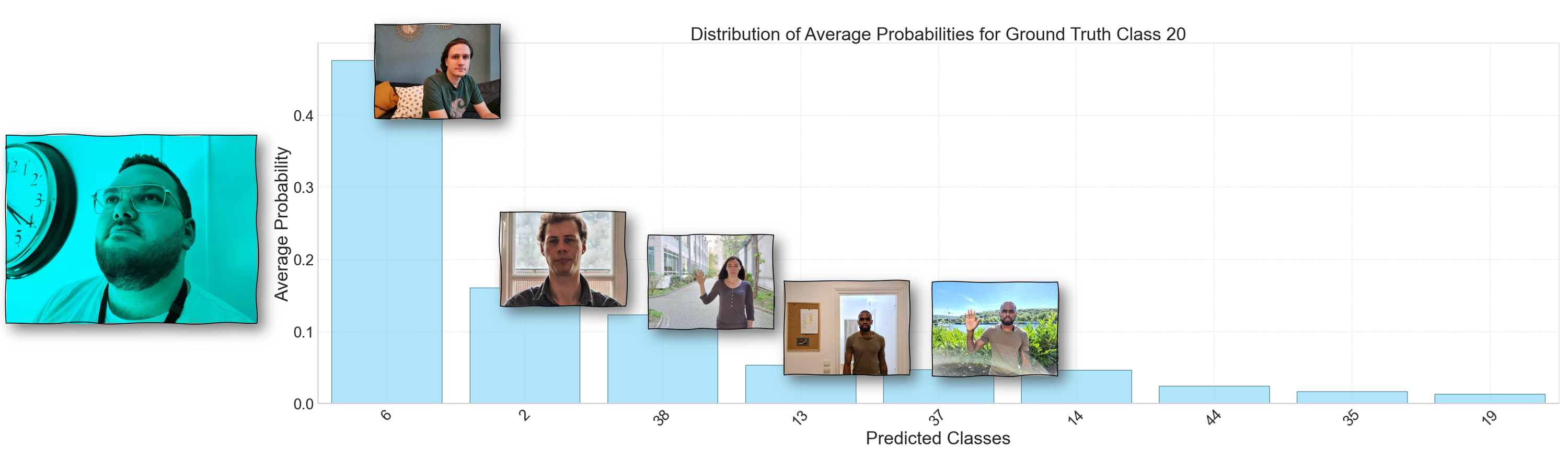} \\
        \includegraphics[width=\columnwidth]{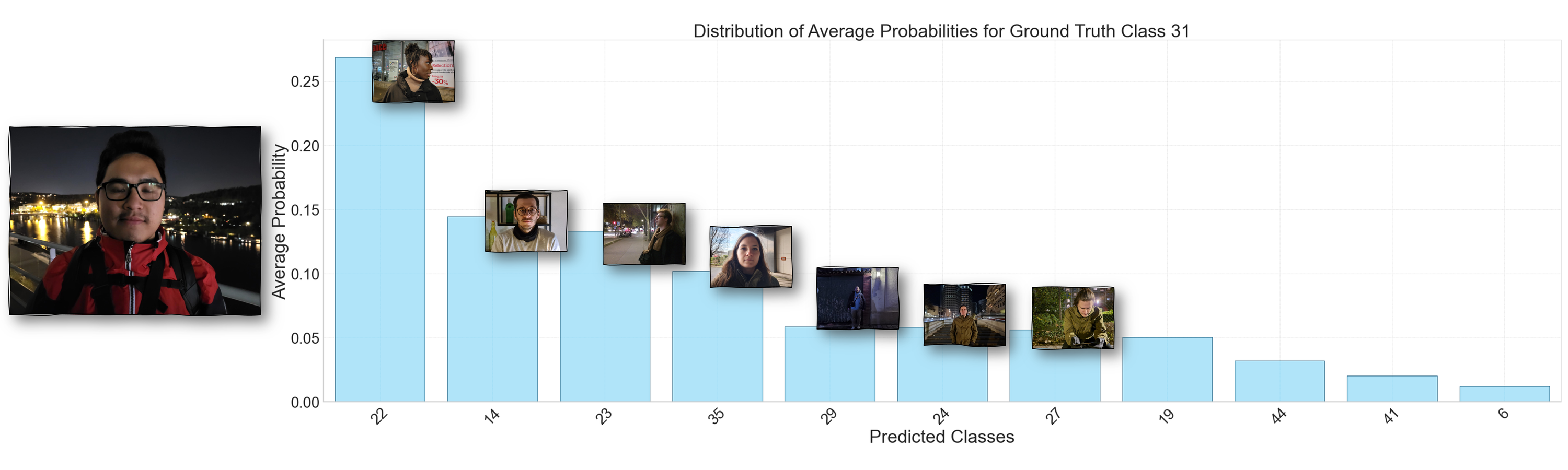} &
        \includegraphics[width=\columnwidth]{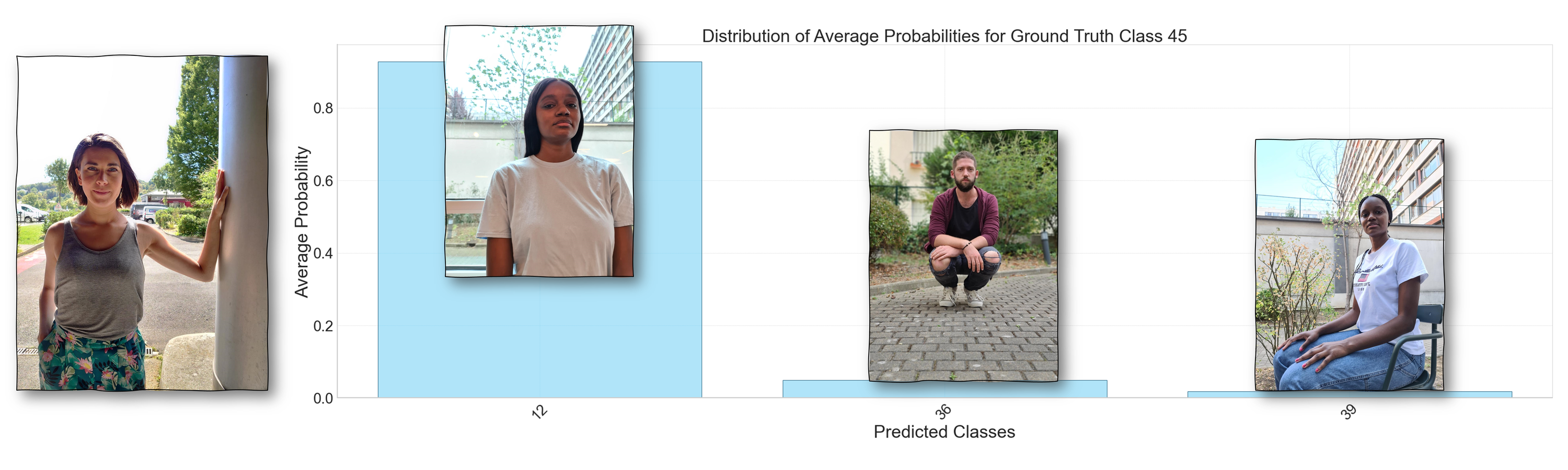} \\
    \end{tabular}
    \caption{Histograms showing the classification distribution across training scenes for various unseen testing conditions. The same testing scene can be projected to multiple training scenes with similar features, showcasing the necessity to consider multiple scenes for inference on new conditions.}
    \label{fig:histograms}
\end{figure*}

The results presented in Tab. \ref{tab:benchmarks} and Fig. \ref{fig:boxplots} offer several significant insights. A deep look into the performance metrics reveals that FHIQA is consistently competitive across the different attributes. It stands out and outperforms other models for ``Overall", demonstrating its comprehensive assessment capabilities and semantic understanding. Specifically, for the ``Overall" attribute, we utilize the entire image, in contrast to other attributes where the evaluation is confined solely to the face region. Therefore, where semantic understanding matters the most, FHIQA proves that it can adapt the knowledge from the training set to generalize to new conditions. Furthermore, when looking at the three attributes altogether, our model is best in 7 out of 15 cases and second best in 6 others. On the one hand, where other HyperIQA variants perform consistently worse, FHIQA stands out as a better solution for generalization, even when compared to MUSIQ, which is pre-trained on IQA datasets. On the other hand, MUSIQ variants also display compelling performances. The PaQ-2-PiQ pre-trained variant dominates in ``Exposure". In contrast, DB-CNN doesn't fare as well as the other models.
In summary, while different models exhibit strengths in specific areas, FHIQA, with no prior IQA pre-training, stands out, emphasizing its generalization and contextual assessment capabilities across a diverse range of attributes and conditions.

\subsection{Understanding new content}
Fig. \ref{fig:histograms} illustrates the distribution of FHIQA's predicted classes for unfamiliar conditions. These distributions highlight the significance of semantic discovery when generalizing an IQA measure for unseen content. For instance, scene 17 depicts a man at his computer desk in a lowlight setting. The model projects this scene onto related contexts like desk scenes, library settings, and other lowlight scenarios. FHIQA leverages these diverse classification distributions to derive a quality assessment that integrates various content perspectives, rather than restricting itself to a singular scene or content. This approach recognizes the pivotal role of scene semantics in determining image quality. The model's robust performance across all attributes —particularly its standout results on ``Overall"—  points to the need for future IQA models to offer content-specific evaluations that are both precise and nuanced.

\section{Conclusion}
In this paper, we have introduced Full-HyperIQA (FHIQA), a novel BIQA method focused on scene generalization. We hypothesize that the information needed for generalization is naturally encoded in the class prediction weights and that the quality of unfamiliar conditions can be extracted based on similar conditions in the training set. Our model obtains competitive or top performance on all attributes of PIQ23, demonstrating the effectiveness of our approach. This paper underscores the pivotal role of semantic understanding in achieving effective generalization for image quality assessment, particularly in portrait scenarios. We advocate for the IQA community to move towards content-specific evaluations, especially in the field of smartphone photography, with the emergence of AI-driven image enhancement, which challenges the traditional IQA methodologies. FHIQA represents a step in this direction, setting the stage for future IQA models that can adapt to varied scenarios and quality criteria.

\printbibliography

\end{document}